\documentclass[runningheads]{llncs}
\usepackage{graphicx}
\usepackage{comment}
\usepackage{amsmath,amssymb} 
\usepackage{color}
\usepackage{algorithm}
\usepackage{algorithmic}
\usepackage{multirow}
\usepackage{xcolor}
\usepackage{booktabs}

\begin{document}
\pagestyle{headings}
\mainmatter
\def\ECCVSubNumber{4985}  

\title{AFAT: Adaptive Failure-Aware Tracker for Robust Visual Object Tracking} 

\titlerunning{Abbreviated paper title}
%
\author{Tianyang Xu$^{1}$  ~~Zhen-Hua Feng$^1$  ~~Xiao-Jun Wu$^{2}$ ~~Josef Kittler$^1$\\
\institute{Centre for Vision, Speech and Signal Processing (CVSSP), University of Surrey, Guildford, UK\and
School of Internet of Things Engineering, Jiangnan University, Wuxi, China}
\email{\tt\small tianyang.xu@surrey.ac.uk, z.feng@surrey.ac.uk, wu\_xiaojun@jiangnan.edu.cn, j.kittler@surrey.ac.uk}}

\maketitle

\begin{abstract}
Siamese approaches have achieved promising performance in visual object tracking recently. The key to the success of Siamese trackers is to learn appearance-invariant feature embedding functions via pair-wise offline training on large-scale video datasets. However, the Siamese paradigm uses one-shot learning to model the online tracking task, which impedes online adaptation in the tracking process. Additionally, the uncertainty of an online tracking response is not measured, leading to the problem of ignoring potential failures. In this paper, we advocate online adaptation in the tracking stage. To this end, we propose a failure-aware system, realised by a Quality Prediction Network (QPN), based on convolutional and LSTM modules in the decision stage, enabling online reporting of potential tracking failures. Specifically, sequential response maps from previous successive frames as well as current frame are collected to predict the tracking confidence, realising spatio-temporal fusion in the decision level. In addition, we further provide an Adaptive Failure-Aware Tracker (AFAT) by combing the state-of-the-art Siamese trackers with our system. The experimental results obtained on standard benchmarking datasets demonstrate the effectiveness of the proposed failure-aware system and the merits of our AFAT tracker, with outstanding and balanced performance in both accuracy and speed.
\keywords{Visual Object Tracking, Convolutional Neural Network, Long Short-Term Memory, Online Adaptation}
\end{abstract}

\section{Introduction}
Visual object tracking is a fundamental computer vision topic with the aim of automatically predicting the state of a target that is initialised with ground truth only in the first frame.
With the exploration of mathematical modelling techniques, especially deep neural networks, recent advanced tracking approaches focus on employing large labelled video datasets to offline train end-to-end networks.
The performance of these end-to-end deep Siamese neural networks, \textit{e.g.}, SINT~\cite{tao2016siamese}, SiameseFC~\cite{bertinetto2016fully}, SiamRPN~\cite{li2018high}, Siam R-CNN~\cite{voigtlaender2019siam}, has witnessed a continuous improvement with the help of more effective network architectures and more public available labelled video datasets.
One typical property of the Siamese paradigm is that it considers an online tracking task as a pair-wise matching problem, where one-shot learning provides suitable mechanistic explanations.
However, in the tracking stage, Siamese approaches fix the initial template model for the entire video sequences without online adaptation, leading to potential tracking shifts or failures.

In general, pair-wise offline training of Siamese networks can achieve effective feature embedding. However, it overemphasises robust appearance feature extraction for specific target categories in the training dataset~\cite{li2019siamrpn++}.
Therefore, the learned model enables high tracking accuracy in the test videos where the static and dynamic appearance variations are similar to those found in the training stage.
However, videos from practical surveillance and mobile devices are of high diversity with countless object categories, not to mention the corresponding variations.
These videos, exhibiting unpredictable appearance content, raise challenges for one-shot learning approaches.

\begin{figure}[t]
\begin{center}
\includegraphics[trim={0mm 40mm 30mm 0mm},clip,width=1\linewidth]{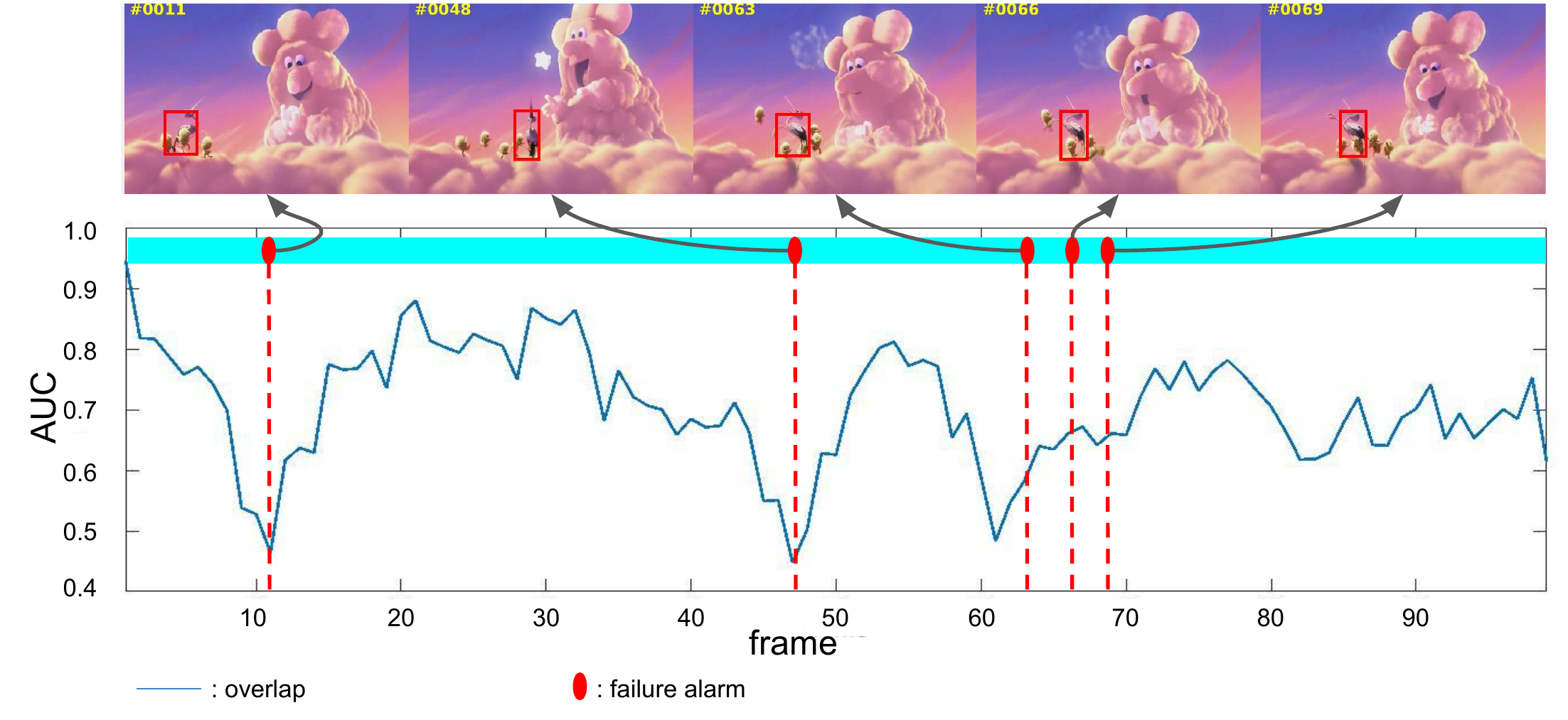}
\end{center}
\caption{Illustration of the proposed adaptive failure-aware tracker. AFAT achieves online quality prediction based on spatio-temporal modelling in the decision level, enabling failure alarm and further correction. AFAT detects 5 failures in the above sequence (\textit{Bird2}), \textit{i.e.}, \#11, \#48, \#63, \#66, \#69, corresponding to severe appearance changes with lower IOU between the tracking result and ground truth. \#11 and \#48 are true positive alarm signals with the overlap falling rapidly.  \#63, \#66 and \#69 are false positive alarm signals as they are close to the previous valley point \#61.
}
\label{illustration}
\end{figure}

To mitigate the above issues, we argue that it is essential to incorporate an online alarm mechanism (failure detection)  in the tracking process.
We propose to integrate an online quality prediction of the tracking results at the decision level, as shown in Fig.~\ref{illustration}.
The response maps from recent successive frames are used as an input to assess the uncertainty of the tracking result of the current frame.
It should be noted that the use of response maps has been widely studied in visual tracking for a variety of goals, ranging from basic target localisation to additional tracking bonus, \textit{e.g.}, quality assessment~\cite{wang2017large,lukezic2017discriminative,bhat2018unveiling}, re-detection condition~\cite{ma2015long,yan2019skimming,xu2019learningl} and update requirement~\cite{danelljan2019atom,bhat2019learning}.
However, existing approaches only use hand-crafted functions or a pre-defined threshold to tap the power of response maps, neglecting their diverse variations in complex tracking situations.
In contrast, we employ convolutional and Long Short-Term Memory (LSTM) modules to extract spatio-temporal fusion features from online generated response maps to perform failure flagging.
Specifically, the response maps are considered as high-level representations, reflecting the probability of each position becoming the target centre.
We show that by extracting implicit decision information from multi-frame response maps via convolutional and LSTM modules enables a reliable and robust prediction of  potential tracking failures in a spatio-temporal perspective.

In the data collection stage, we perform online tracking on random video segments from video datasets (TrackingNet~\cite{muller2018trackingnet} in our experiment).
The response maps and corresponding quality states are collected and stored as training samples for our Quality Prediction Network (QPN).
We assign the quality state for each frame based on the Intersection Over Union (IOU) measure between the predicted bounding box and the corresponding ground truth.
In the offline training phase, the labelled response map sequences are fed into QPN with a cross entropy loss.
As temporal order and 2D response scores are jointly considered in network training, adaptive prediction of the tracking quality can be achieved by our QPN.
A forward pass of QPN takes only 1.2$\sim$1.5 ms, which can be easily integrated with existing methods.
Further more, we combine QPN with the state-of-the-art Siamese trackers for robust online visual object tracking, providing an effective failure detection and further performance boosting.

The constructed Adaptive Failure-Aware Tracker (AFAT) enables an adaptive model selection in the online tracking process, comprehensively predicting the current tracking difficulty and assigning the tracker with a suitable model capacity.

The main contributions of the proposed AFAT method include:
\begin{itemize}
\item A new quality prediction network (QPN) for generating online tracking alarm signal, and predicting potential failures in the tracking process, achieving spatio-temporal responses fusion at the decision level. The temporal order and spatial appearance of multi-frame response maps are jointly considered to predict the reliability of the tracking results, providing additional functionality to existing one-shot learning algorithms.

\item A new adaptive failure-aware tracker equipped with advanced Siamese trackers and QPN, enabling synchronous adaptive model selection in online object tracking, with a perceptual model configuration and tracking correction.

\item A comprehensive evaluation of our AFAT method on a number of well-known benchmarking datasets, including OTB2015~\cite{wu2015object}, UAV123~\cite{mueller2016benchmark}, LaSOT~\cite{fan2019lasot}, VOT2016~\cite{Kristan2016The}, VOT2018~\cite{kristan2018sixth} and VOT2019~\cite{kristan2019seventh}.
The results demonstrate the superior performance of the proposed AFAT method over the state-of-the-art trackers.
\end{itemize}

\noindent The rest of this paper is organised as follows:
In Section~\ref{background}, we provide the motivation for introducing a failure-aware mechanism into tracking and discuss the related studies in recent development.
A detailed construction and training scheme of the proposed quality prediction network are presented in Section~\ref{qpnsection} and furnished with an efficient adaptive failure-aware tracker in Section~\ref{afat}.
The experimental results and the corresponding analysis are reported in Section~\ref{experiment}.
Last, the conclusions are drawn in Section~\ref{conclusion}.

\section{Background}\label{background}
\subsection{Advanced Tracking models}
Recent studies in visual object tracking focus on modelling the task as online discriminative learning or offline pair-wise matching network, both achieving continuously developed performance in standard benchmarks~\cite{wu2015object,liang2015encoding,mueller2016benchmark,fan2019lasot,muller2018trackingnet} and competitions~\cite{kristan2018sixth,kristan2019seventh}.

Online discriminative learning was designed to distinguish the target region from its surroundings via online training and updating a classifier (or regressor).
To alleviate the computational burden for online learning, Discriminative Correlation Filter (DCF) based approaches have extensively studied in recent years, with advanced efficiency and accuracy contributed by the circulant structure~\cite{henriques2012exploiting,henriques2014high,xu2019accelerated}.
Meanwhile, DCF-based methods address the sample sufficiency issue in online discriminative learning by augmented circulant samples.
To optimise the estimate filters with more intuitive requirements and desired constraints, further attempts have been performed in the DCF paradigm, \textit{e.g.}, spatial regularisation for boundary effect~\cite{danelljan2015learning,lukezic2017discriminative,kiani2017learning,xu2018non}, feature selection for input redundancy~\cite{xu2019learning,xu2019joint}, temporal sampling space for filter degeneration~\cite{danelljan2016adaptive,danelljan2016beyond,danelljan2017eco}, and multi-response fusion for extended representation~\cite{bertinetto2016staple,bhat2018unveiling}.

Offline pair-wise matching network benefits from the design of the Siamese network architecture.
Specifically, a Siamese network maps the image pairs, \textit{i.e.} template and instance, into a latent feature space, where the target regions are of salient similarity against practical appearance variations.
After feature embedding, SINT~\cite{tao2016siamese} compares the scores of candidates for final decision making, and GOTURN~\cite{held2016learning} performs bounding box localisation using FC layers.
To avoid multi-candidate calculation and FC layer learning, cross correlation is directly used for response calculation in SiameseFC~\cite{bertinetto2016fully}, realising a fully convolutional structure.
Besides the effective capability in centre localisation, the accuracy of the final bounding box is also essential for advanced tracking methods.
To this end, recent efforts have been made on fusing bounding box refinement techniques in the offline training stage.
SiamRPN utilises the Region Proposal Network (RPN) for joint high-quality foreground-background classification and bounding box regression learning~\cite{li2018high}.
While ATOM performs IOU maximisation for fine-grained overlap precision~\cite{danelljan2019atom}.
In addition, to improve the exploitation from diverse multi-layer features and Siamese layers, SiamRPN++~\cite{li2019siamrpn++} and SiamDW~\cite{zhang2019deeper} are proposed, which demonstrate promising performance for visual object tracking.

Despite the success in designing powerful tracking models, the significance of online tracking quality prediction and the corresponding countermeasures are underestimated by the community.
In other words, the pattern of the tracking calculation process is predefined by a specified mathematical model, which always assumes specific distributions of the target appearance or motion trajectory.
We argue the necessity of online predicting the tracking quality and performing effective intervention and correction.

\subsection{Online tracking adaptation}
Though continuous improvements have been achieved in developing diverse mathematical formulations in visual tracking, a practical tracking task can not be exactly modelled by existing methodologies accurately.
To mitigate this issue, online controlling strategies are necessary to balance the characteristics of the theories reflecting simplified assumptions about the universe of target tracking and their validity in challenging video sequences with unpredictable appearance variations.

To achieve the above goal, many tracking approaches propose to analyse the reliability of the current result for potential update and re-detection.
To alleviate tracking shift caused by radical update, ATOM~\cite{danelljan2019atom} and Dimp~\cite{bhat2019learning} adjust the learning rate based on the response score.
TLD~\cite{kalal2011tracking} utilised an independent classifier to re-detect the target with a distance threshold-based failure detector.
Similar threshold-based quality measurement is widely used in recent online discriminative learning and Siamese trackers, \textit{e.g.},
LCT~\cite{ma2015long}, SPLT~\cite{yan2019skimming} and Siamese-LT~\cite{li2019siamrpn++}, performing re-detection when the response score is under a threshold.
Besides, quality prediction also achieved wide attention for online adaptation in visual tracking.
LMCF~\cite{wang2017large} utilised the response map for multi-modal target detection to reduce the impact from similar appearance.
Then, a quality function is designed for high-confidence update, suppressing the contaminated samples.
Other attempts in quality prediction focus on adaptive multi-response fusion.
CSR-DCF~\cite{lukezic2017discriminative} assigns more reliability to the single modal response channels and UPDT~\cite{bhat2018unveiling} simultaneously constrains the fusion response to be narrow and uni-modal distributed.

However, the above processing techniques only use a simple threshold or hand-crafted function to measure the quality of the current tracking result, which are unable to satisfy the practical tracking challenges.
Therefore, we propose to employ neural network to predict the tracking quality.
To better extract the potential information from response maps, we construct a spatio-temporal quality prediction network (QPN) with convolutional and LSTM modules.
Different form existing measurement, we collect the response maps from successive multiple frames to enhance the sequential order.
By training from the generated labelled data, QPN enables adaptive classification between successful and failed result in the decision level.

\subsection{Meta-learning for tracking}
From the perspective of learning framework, Siamese trackers belong to one-shot learning category, using meta-learning formulation to learn the best embedding network that highlights the target region while suppresses the background.
In this paradigm, Learnet~\cite{bertinetto2016learning} is the seminal work using meta-learning to generate parameters for visual tracking.
Besides predicting the tracking parameters, Meta-tracker~\cite{park2018meta} proposed to jointly learn the learning rate.
While UpdateNet~\cite{zhang2019learning} proposed to replace the handcrafted update function with a method which learns to update.
In addition, MLT~\cite{choi2019deep} utilised a meta-learner network to provide the Siamese network with new appearance of the target.
To explore the potential power of meta-learning, we propose to train a quality prediction network for a given tracker.



\section{Quality prediction network}
\label{qpnsection}
\begin{figure}[t]
\begin{center}
\includegraphics[trim={0mm 0mm 40mm 0mm},clip,width=1\linewidth]{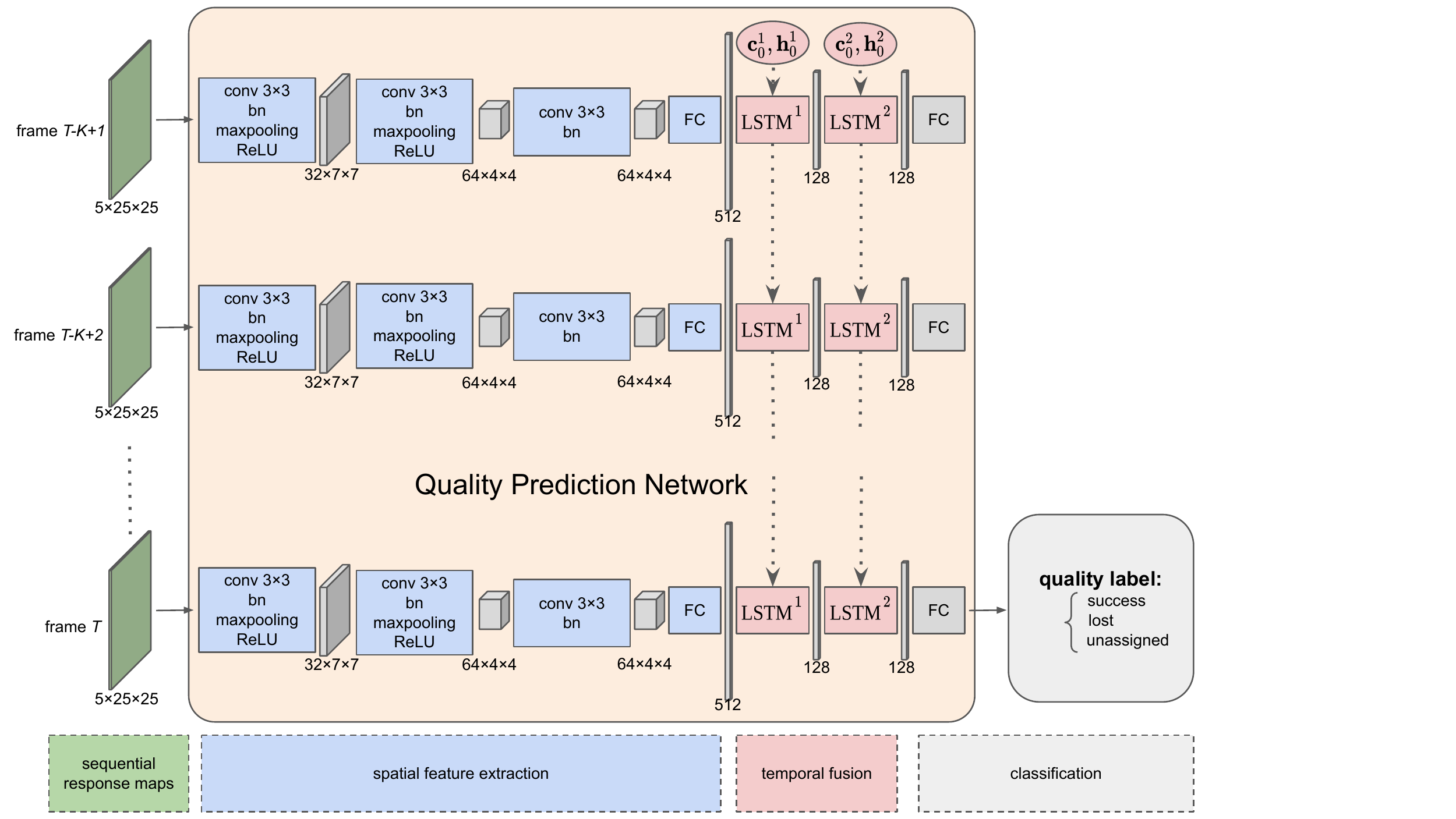}
\end{center}
\caption{The proposed quality prediction network (QPN).}
\label{framework}
\end{figure}
In this section, we present the proposed Quality Prediction Network (QPN).
As shown in Fig.~\ref{framework}, the proposed QPN consists of convolutional and Long Short-Term Memory (LSTM) modules.
Sequential response maps and the corresponding quality labels are used to train QPN in an end-to-end fashion.
QPN utilises the current and previous temporal information to provide quality feedback, \textit{i.e.}, whether the current tracking result is successful or not.
The forward pass of QPN can be formulated as:
\begin{equation}
\label{formulation}
q_t=\textrm{QPN}\left(\mathbf{f}_{t-K+1},\mathbf{f}_{t-K+2},\ldots,\mathbf{f}_{t}\right),
\end{equation}
where $\mathbf{f}_i\in\mathbb{R}^{C\times N\times N}$ is the response maps of the $i$-th frame. We select SiameseRPN++~\cite{li2019siamrpn++} (with mobilev2 backbone) as our base tracker, which can produce the response map of $5\times 25\times 25$ size in each frame, where $5$ and $25\times 25$ are corresponding to the anchor number and spatial resolution, respectively. $K$ is the sequential length for our spatio-temporal inference. We set $K=20$ in our experiment.
We describe the construction of the QPN with more details as follows.

\subsection{Spatial feature extraction}
Different from raw image patches, we use response maps as the input data for our QPN.
Response maps are output by an online tracker, reflecting the decision-making ability of the model in the current tracking situation.
Similar to an image, considering the spatial distribution, we propose to use CNN for basic spatial feature extraction from response maps.
Based on existing studies and early analysis, we have the following observations: (1) an ideal response map should be uni-modal with narrow peak; and (2) a tracking failure always occurs with clutters in the response map.
To make the convolutional network suitable for our response input, we circularly shift the response map and scatter the maximal peak into the four corners, enhancing the context information.
Then three convolutional layers and one FC layer are employed to extract the representations $\phi\left(\mathbf{f}_i\right)$ for each frame $i$ in the response map sequence, where $\phi$ denotes the forward pass of the CNN for spatial feature extraction.

\subsection{Temporal feature fusion}
Existing quality measurements in advanced tracking approaches only consider the response maps obtained by the current frame, neglecting the previous temporal information.
In general, a basic assumption in visual tracking is that the target always smoothly changes in appearance and position between adjacent frames. Therefore, we argue the significance of exploiting temporal ordered data for tracking quality prediction.
To this aim, we propose to fuse the sequential response maps from recent successive frames for quality prediction using two LSTM networks~\cite{gers1999learning}.
The extracted spatial features $\mathbf{\Phi}= \left\{\phi\left(\mathbf{f}_{t-K+1}\right),\phi\left(\mathbf{f}_{t-K+2}\right),\ldots,\phi\left(\mathbf{f}_{t}\right)\right\}$ are fed into the two LSTM modules to predict the quality of the tracking result in frame $t$, $q_t=g\left(\mathbf{\Phi}\right)$.

\subsection{End-to-end training of QPN}
To generate sequential response maps for training our QPN, we run visual tracking using our base tracker (SiameseRPN++, with mobilev2 backbone) on TrackingNet~\cite{muller2018trackingnet}.
There are 30132 labelled video sequences in TrackingNet, and we select the first 20000 videos to generate the training data, and select the rest 10132 videos for validation.
Specifically, for each piece of data, we collected the response maps via running the base tracker on successive 20 frames from a randomly selected video.
In addition, we label the response maps based on the IOU between the predicted bounding box and the corresponding ground truth for each frame,
\begin{equation}
\label{label}
y_i=
\left\{
\begin{aligned}
&success,  &\textrm{if}\ IOU>0.5,\\
&lost,  &\textrm{if}\ IOU<0.1,\\
&\textrm{unassigned}, &\textrm{others}.
\end{aligned}
\right.
\end{equation}
The detailed number of our generated training samples is list in Table~\ref{trackingnet}.

\begin{table}[tb]
\centering
\caption{Data generation for QPN}
\label{trackingnet}
\begin{tabular}{lccccc}
\hline &  sequences & total frames & \multicolumn{3}{c}{labelled frames}\\
& & & success & lost & other\\
\hline
Training set & 50000 & 1000000 & 937881 & 14494 & 47625\\
Validation set & 20000 & 400000 & 375604 & 5429 & 18967\\
\hline
\end{tabular}
\end{table}

\begin{figure}[t]
\begin{center}
\includegraphics[trim={35mm 92mm 40mm 80mm},clip,width=0.32\linewidth]{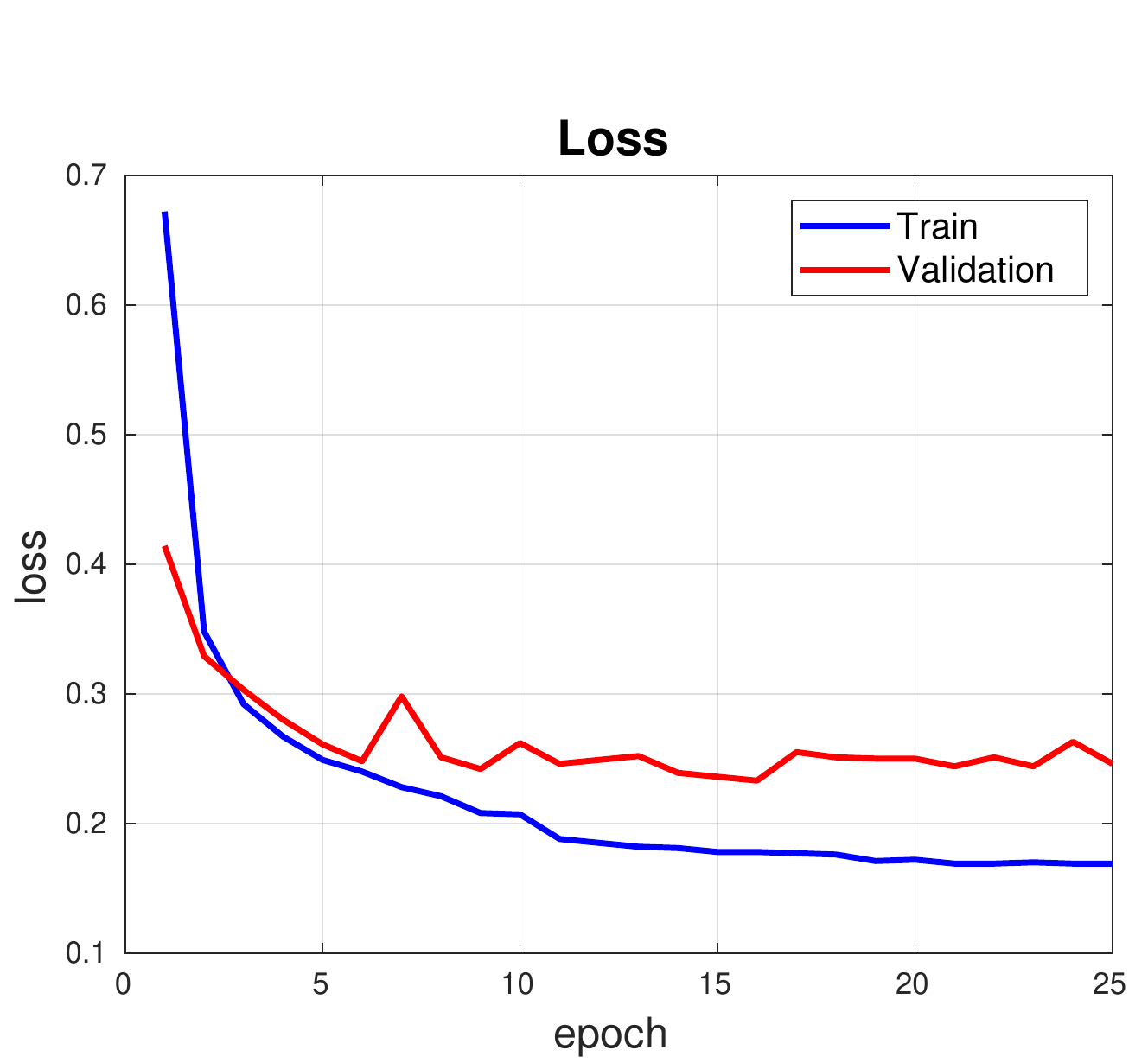}
\includegraphics[trim={35mm 92mm 40mm 80mm},clip,width=0.32\linewidth]{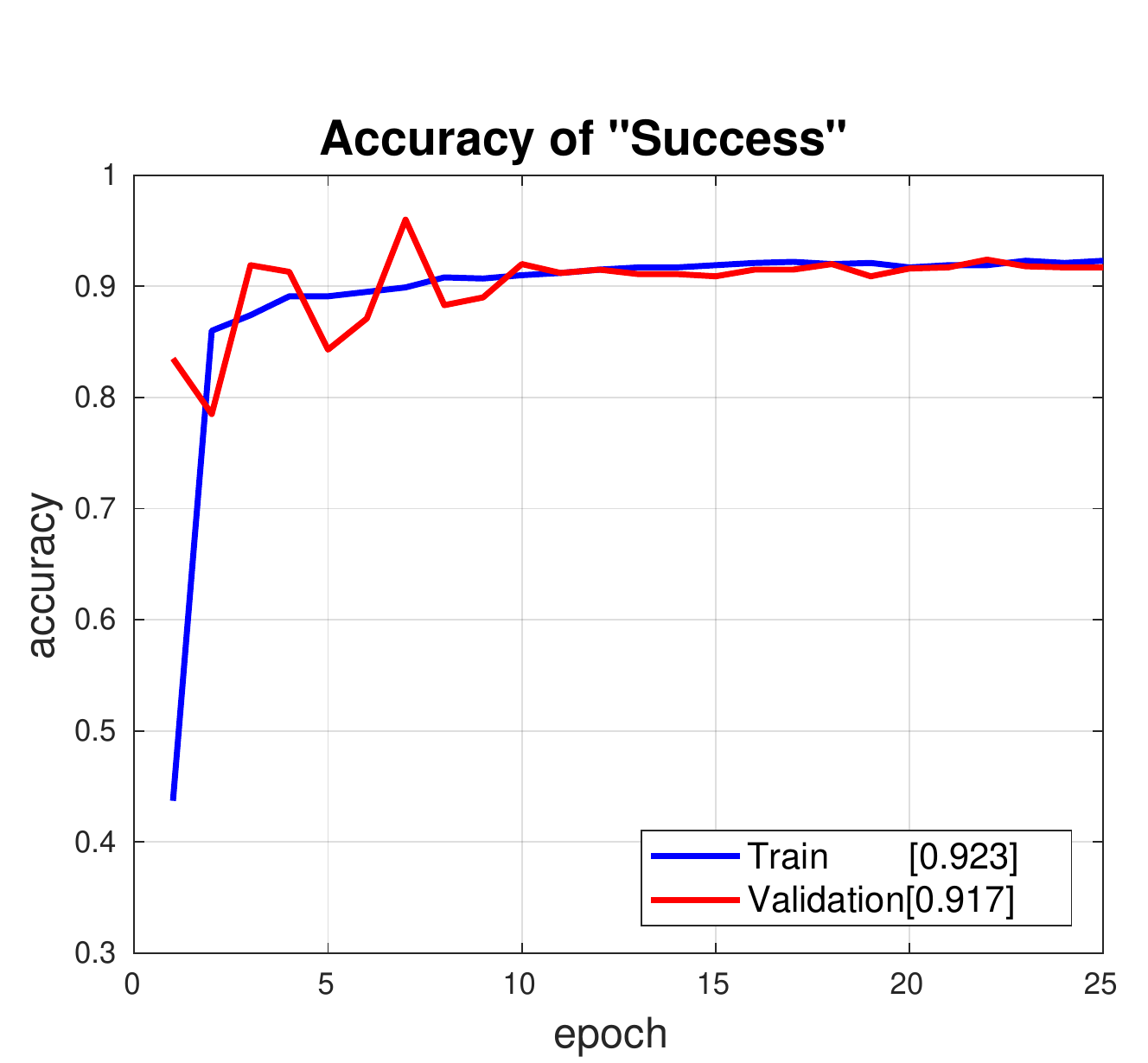}
\includegraphics[trim={35mm 92mm 40mm 80mm},clip,width=0.32\linewidth]{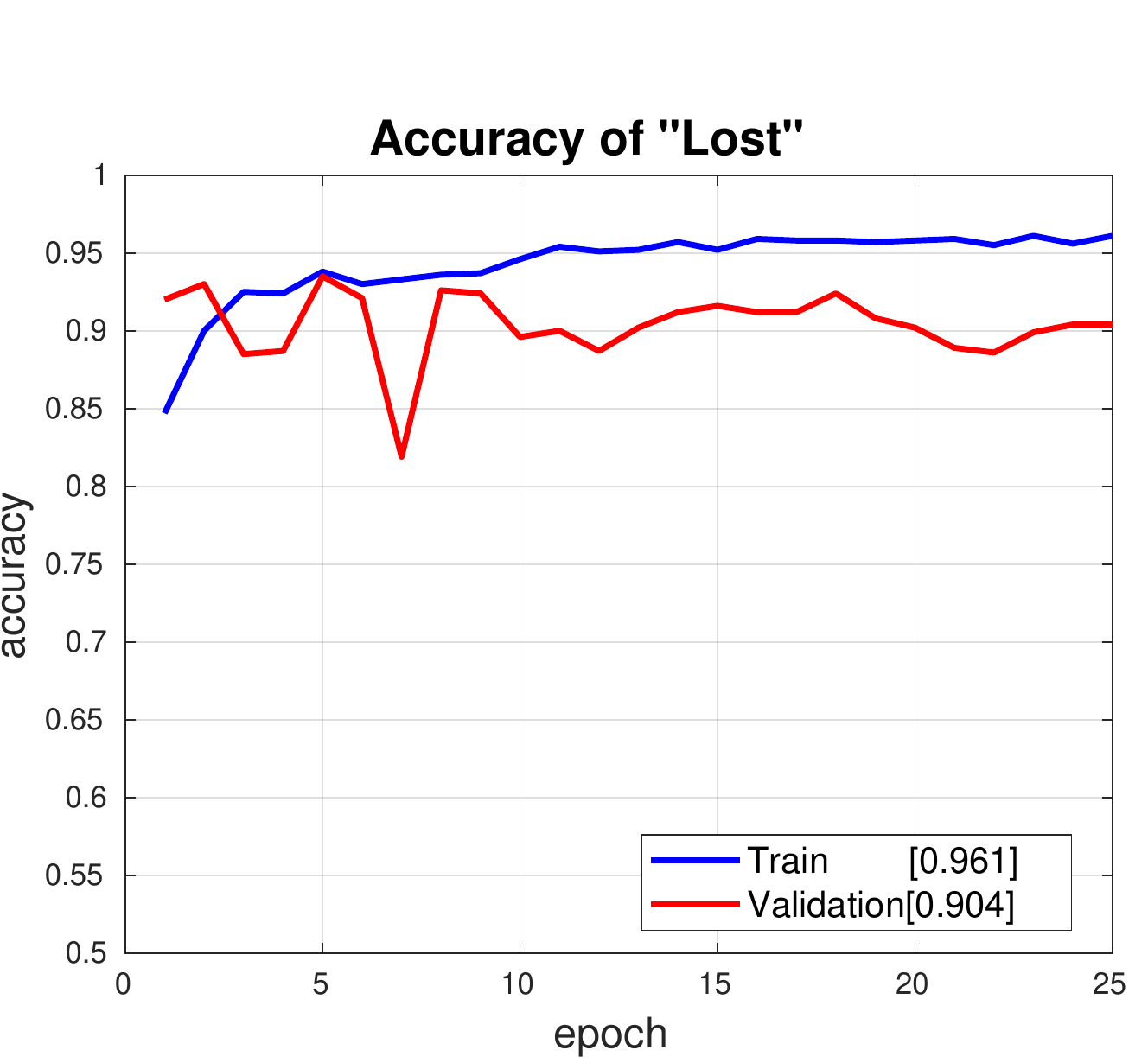}
\end{center}
\caption{Illustration of loss and accuracy in the QPN training phase.}
\label{qpn}
\end{figure}

During the training stage of QPN, sequential response maps $\left(\mathbf{f}_i,y_i\right)$ are fed into the network, where $\mathbf{f}_i\in\mathbb{R}^{20\times 5\times 25\times 25}$ and $y_i\in\mathbb{R}$.
We train QPN from scratch using the Stochastic Gradient Descent (SGD) optimiser.
For network optimisation, the Cross-entropy loss function is used in our design.
As shown in Table.~\ref{trackingnet}, the training data is imbalanced between the \textit{success} and \textit{lost} categories.
To balance the volume between these two types of training data, we set the loss weight as 0.002 and 1 for \textit{success} and \textit{lost}, respectively.
The hidden state and cell state in the two LSTM networks are initialised with zeros for each forward pass.
We train the network for 25 epochs in total.
The learning rate is decreased from 0.01 to 0.001 since the 16-th epoch.
We report the detailed loss and accuracy in each epoch in Fig.~\ref{qpn}.
During the training stage, QPN converges with the accuracy of $0.917$ and $0.904$ on the validation set for \textit{success} and \textit{lost}, respectively.
This demonstrates the effectiveness of the proposed QPN method in failure detection for an online object tracker.
In addition, a forward pass of QPN takes only 1.2$\sim$1.5 ms, which can be easily integrated with existing methods.

\section{Adaptive failure-aware tracker}\label{afat}
We propose the Adaptive Failure-Aware Tracker (AFAT) based on the SiameseRPN++ tracker~\cite{li2019siamrpn++} and the proposed QPN in Section~\ref{qpnsection}. The tracking framework is summarised in Algorithm~\ref{alg}.
We directly use two backbones, \textit{i.e.}, mobilev2 and resnet50, in our AFAT.
We employ $\textrm{SiameseRPN++}_\textrm{mobilev2}$ as the base tracker and $\textrm{SiameseRPN++}_\textrm{resnet50}$ as the correction tracker.
Without bells and whistles, the default parameters from the original SiameseRPN++ are used in the proposed AFAT method.

\begin{algorithm}[t]
\begin{algorithmic}
\vspace{0.03in}
\STATE
\textbf{Initialisation:}
Initialise $\textrm{SiameseRPN++}_\textrm{mobilev2}$ and $\textrm{SiameseRPN++}_\textrm{resnet50}$ with the ground truth in the initial frame.

\STATE
\textbf{Input:}
Image frame $\textbf{I}_t$, model $\textrm{SiameseRPN++}_\textrm{mobilev2}$ and $\textrm{SiameseRPN++}_\textrm{resnet50}$, response list $\textbf{I}$,target centre coordinate $p_{t-1}$ and scale size $w\times h$ from frame $t-1$;

\textbf{Tracking:}

\STATE 1. Extract search windows from $\textbf{I}_t$ at $p_{t-1}$;

\STATE 2. Calculate response maps $\textbf{f}$ using $\textrm{SiameseRPN++}_\textrm{mobilev2}$;

\STATE 3. Add $\textbf{f}$ into $\textbf{I}$, if the length of  $\textbf{I}$ exceeds 20, remove the early observations;

\STATE 4. Record current tracking result from model $\textrm{SiameseRPN++}_\textrm{mobilev2}$;

\STATE 5. Forward pass QPN and obtain the quality prediction $q_t$;

\STATE 6. If $q_t = lost$, change current tracking result using model $\textrm{SiameseRPN++}_\textrm{resnet50}$ and empty $\textbf{I}$;

\STATE
\textbf{Output:} Target bounding box (centre coordinate $p_t$ and current scale size $w\times h$).
\vspace{0.03in}
\end{algorithmic}
\caption{AFAT tracking algorithm.}
\label{alg}
\end{algorithm}

\begin{table*}[t]
\footnotesize
\renewcommand{\arraystretch}{1.1}
\caption{Ablation study of the proposed AFAT on VOT2016, VOT2018,  VOT2019, OTB2015, UAV123 and LaSOT.}
\label{aba}
\centering
\begin{tabular}{l|ccc|ccc|ccc}
\toprule[2pt]
& \multicolumn{3}{c|}{\textbf{VOT2016}} & \multicolumn{3}{c|}{\textbf{VOT2018}} & \multicolumn{3}{c}{\textbf{VOT2019}}\\
& EAO & A & FPS & EAO & A & FPS & EAO & A &FPS\\
\hline
\hline
SiamRPN++$_\textrm{mobilev2}$ & 0.455& 0.624&\textbf{96.4} & 0.410& 0.586& \textbf{95.8}& 0.291& 0.579&\textbf{93.0}\\
SiamRPN++$_\textrm{resnet50}$ & 0.464& \textbf{0.642}& 38.7& 0.414& 0.600 & 37.9& 0.287& 0.594& 38.2\\
\textbf{AFAT}  & \textbf{0.482} & \textbf{0.642} &74.0 & \textbf{0.419}& \textbf{0.605} & 70.5& \textbf{0.295}& \textbf{0.599}& 70.9\\
\hline
Failure detections&\multicolumn{3}{c|}{6632}&\multicolumn{3}{c|}{7007}&\multicolumn{3}{c}{6814}\\
Total frames&\multicolumn{3}{c|}{21455}&\multicolumn{3}{c|}{21356}&\multicolumn{3}{c}{19935}\\
Failure detection rate&\multicolumn{3}{c|}{30.9\%}&\multicolumn{3}{c|}{32.8\%}&\multicolumn{3}{c}{34.2\%}\\
\toprule[2pt]
& \multicolumn{3}{c|}{\textbf{OTB2015}} & \multicolumn{3}{c|}{\textbf{UAV123}} & \multicolumn{3}{c}{\textbf{LaSOT}}\\
& AUC & DP & FPS & AUC &DP & FPS & AUC& NP &FPS\\
\hline
\hline
SiamRPN++$_\textrm{mobilev2}$ & 0.658& 0.864&\textbf{93.2} & 0.602& 0.802& \textbf{95.1}& 0.450& 0.525& \textbf{95.2}\\
SiamRPN++$_\textrm{resnet50}$ & \textbf{0.663}& \textbf{0.875}& 38.5& 0.611& 0.804 & 37.9& \textbf{0.496}& 0.571&38.9 \\
\textbf{AFAT}  & \textbf{0.663} & 0.874 & 86.8& \textbf{0.612}& \textbf{0.811} &80.4 & 0.492& \textbf{0.574}& 77.7\\
\hline
Failure detections&\multicolumn{3}{c|}{2881}&\multicolumn{3}{c|}{4513}&\multicolumn{3}{c}{57192}\\
Total frames&\multicolumn{3}{c|}{59035}&\multicolumn{3}{c|}{112578}&\multicolumn{3}{c}{685360}\\
Failure detection rate&\multicolumn{3}{c|}{4.88\%}&\multicolumn{3}{c|}{4.01\%}&\multicolumn{3}{c}{8.34\%}\\
\toprule[2pt]
\end{tabular}
\end{table*}

\begin{figure}[!t]
\begin{center}
\includegraphics[trim={0mm 40mm 45mm 0mm},clip,width=0.49\linewidth]{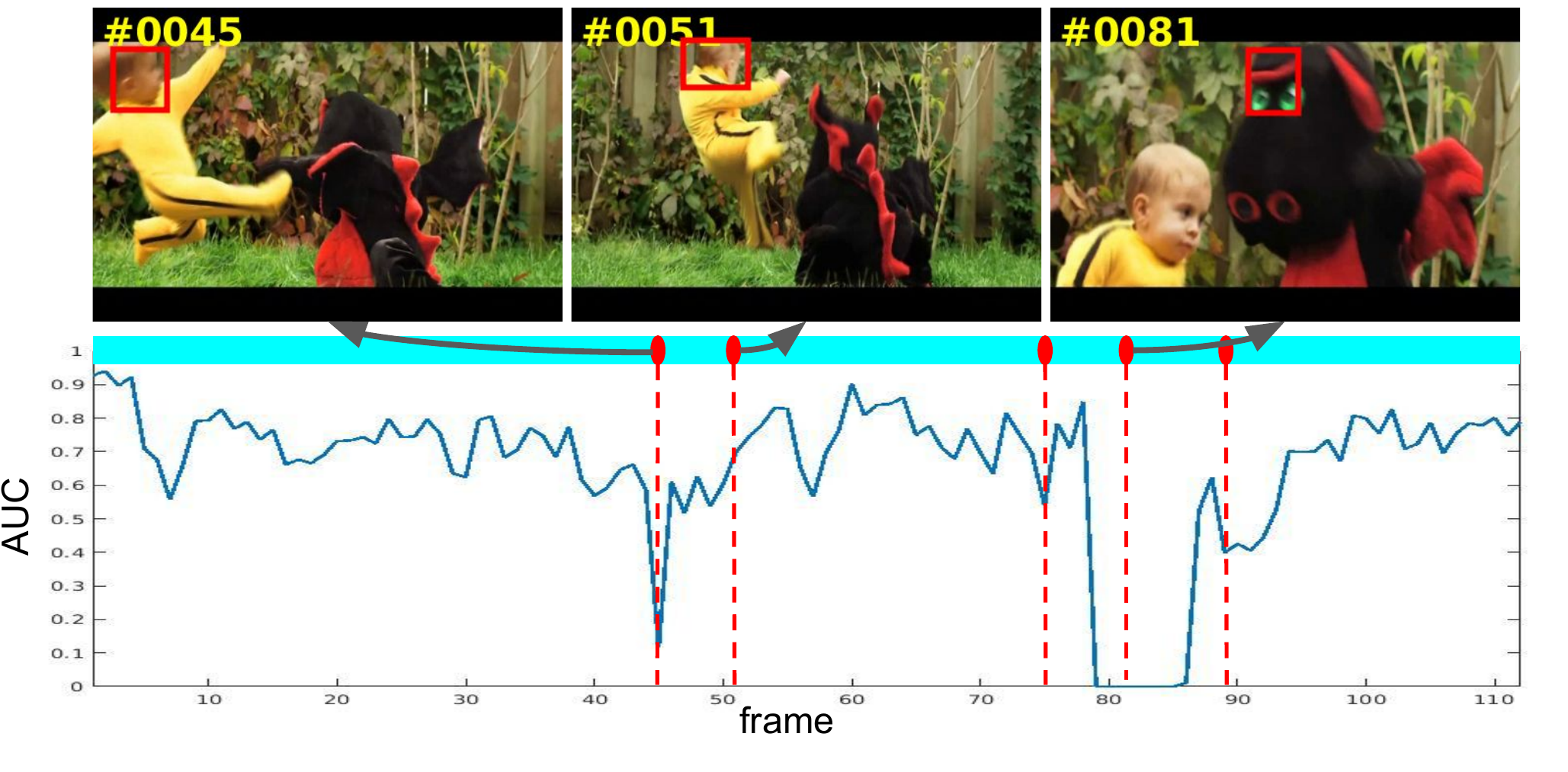}
\includegraphics[trim={0mm 40mm 45mm 0mm},clip,width=0.49\linewidth]{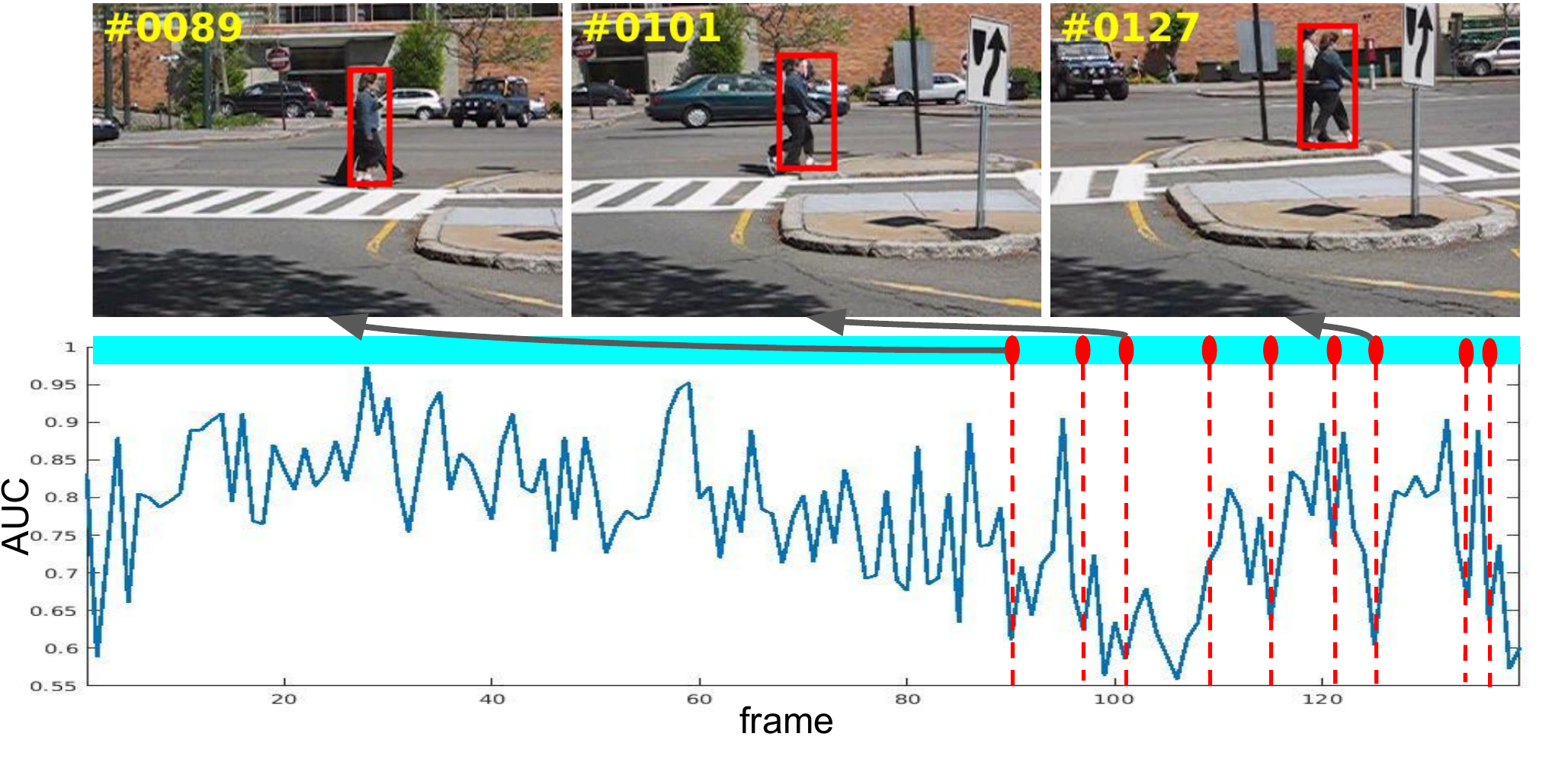}\\
\includegraphics[trim={0mm 40mm 45mm 0mm},clip,width=0.49\linewidth]{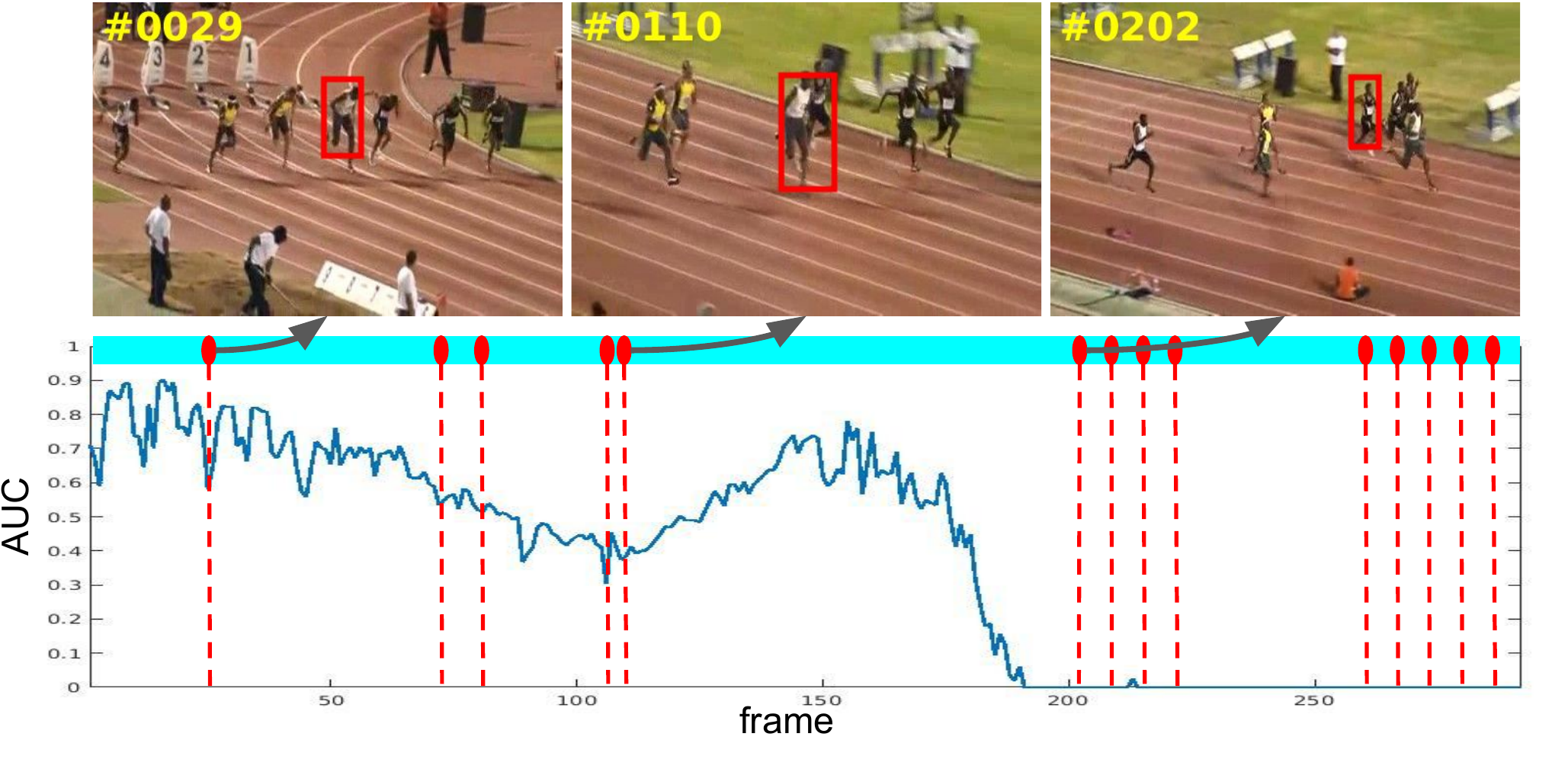}
\includegraphics[trim={0mm 40mm 45mm 0mm},clip,width=0.49\linewidth]{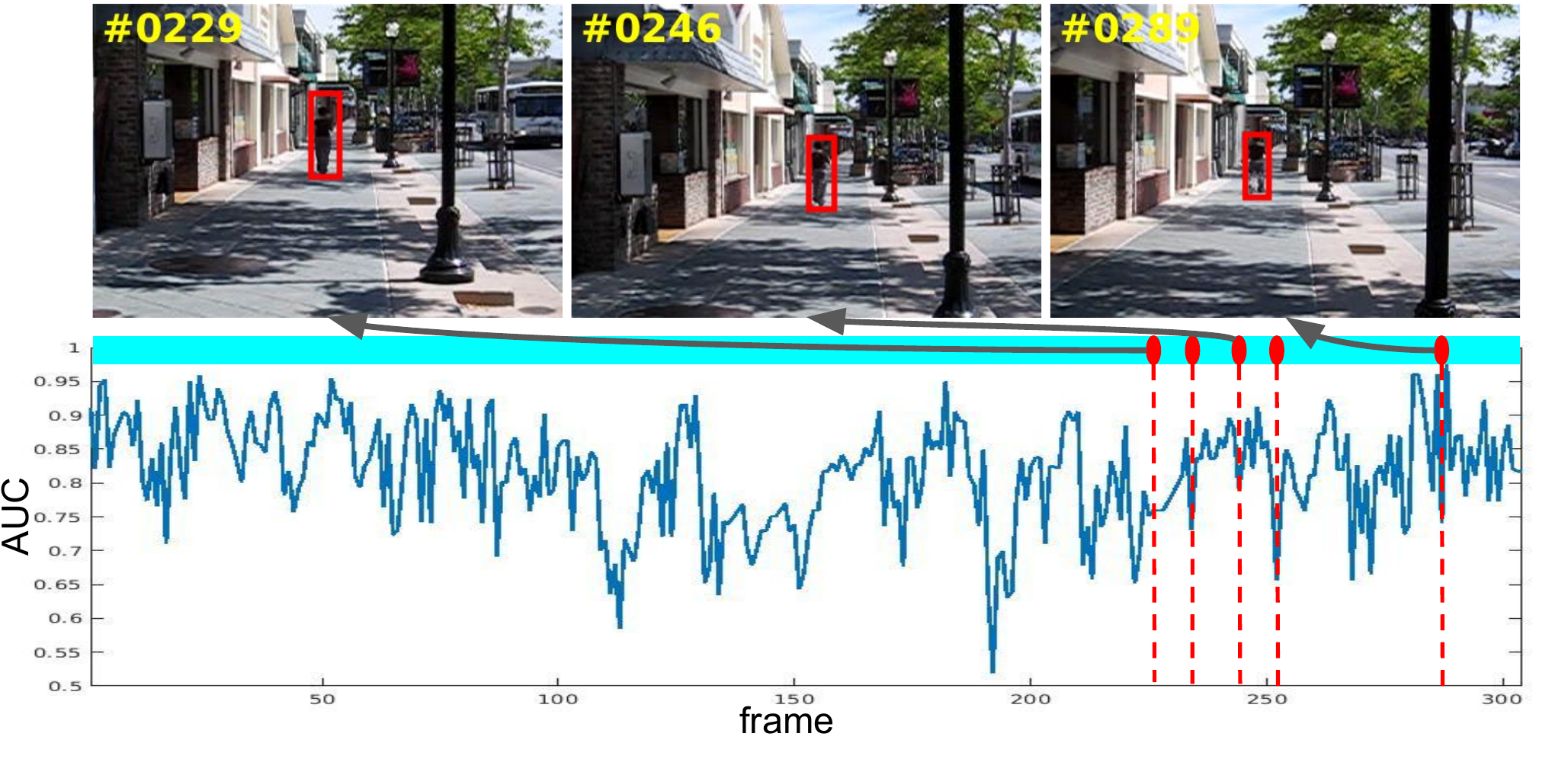}\\
\includegraphics[trim={0mm 40mm 45mm 0mm},clip,width=0.49\linewidth]{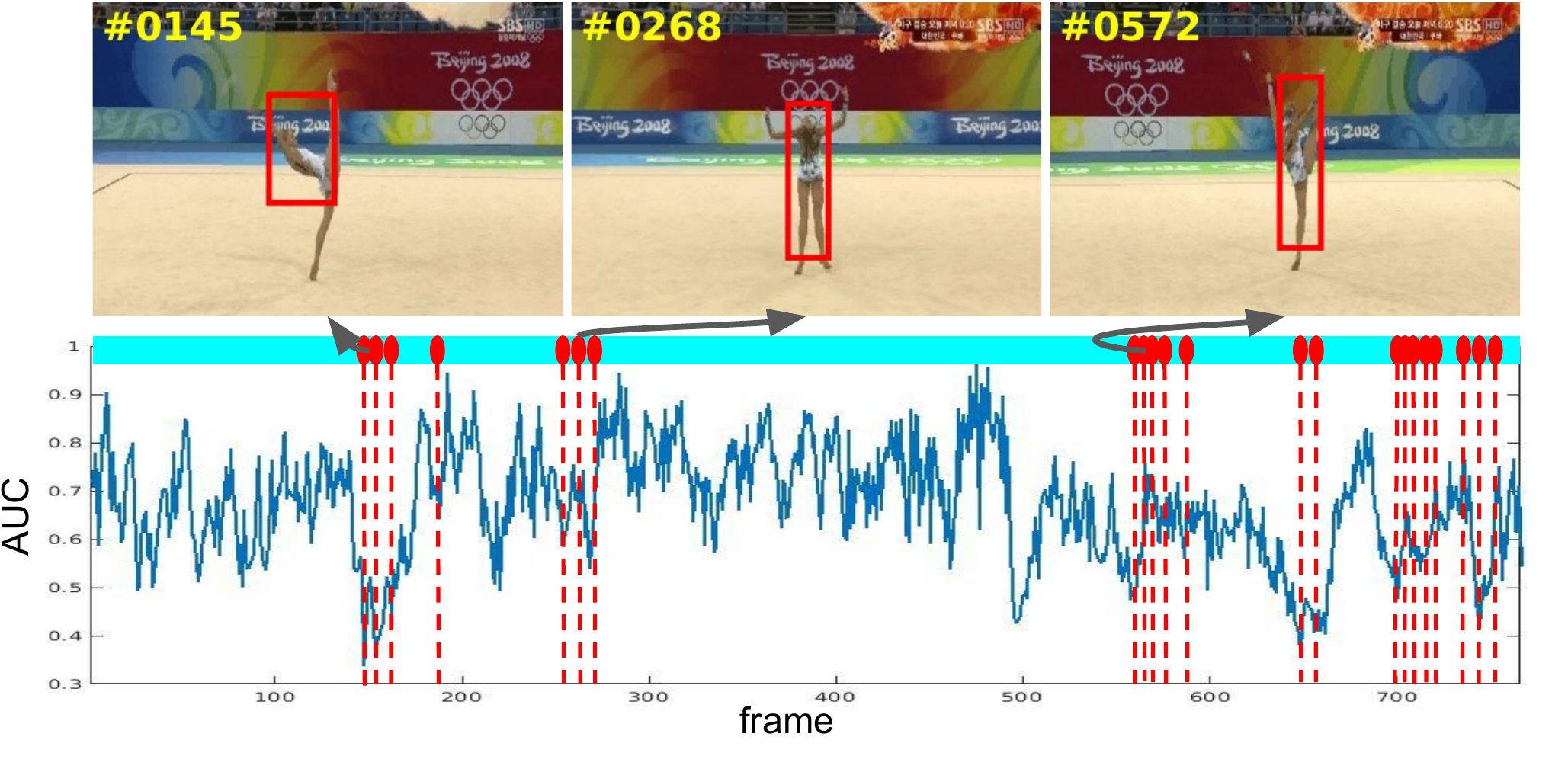}
\includegraphics[trim={0mm 40mm 45mm 0mm},clip,width=0.49\linewidth]{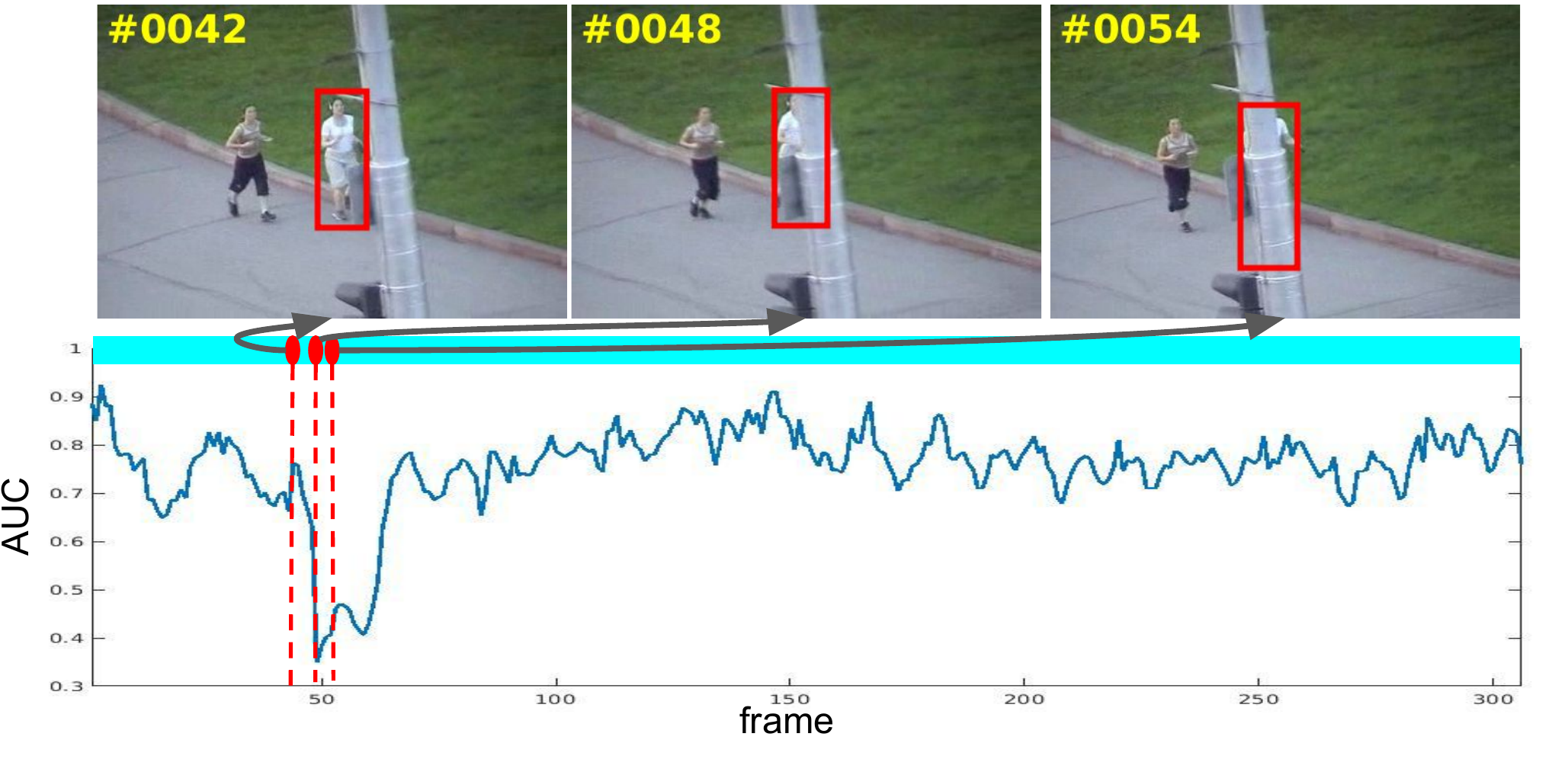}\\
\includegraphics[trim={0mm 45mm 45mm 0mm},clip,width=0.49\linewidth]{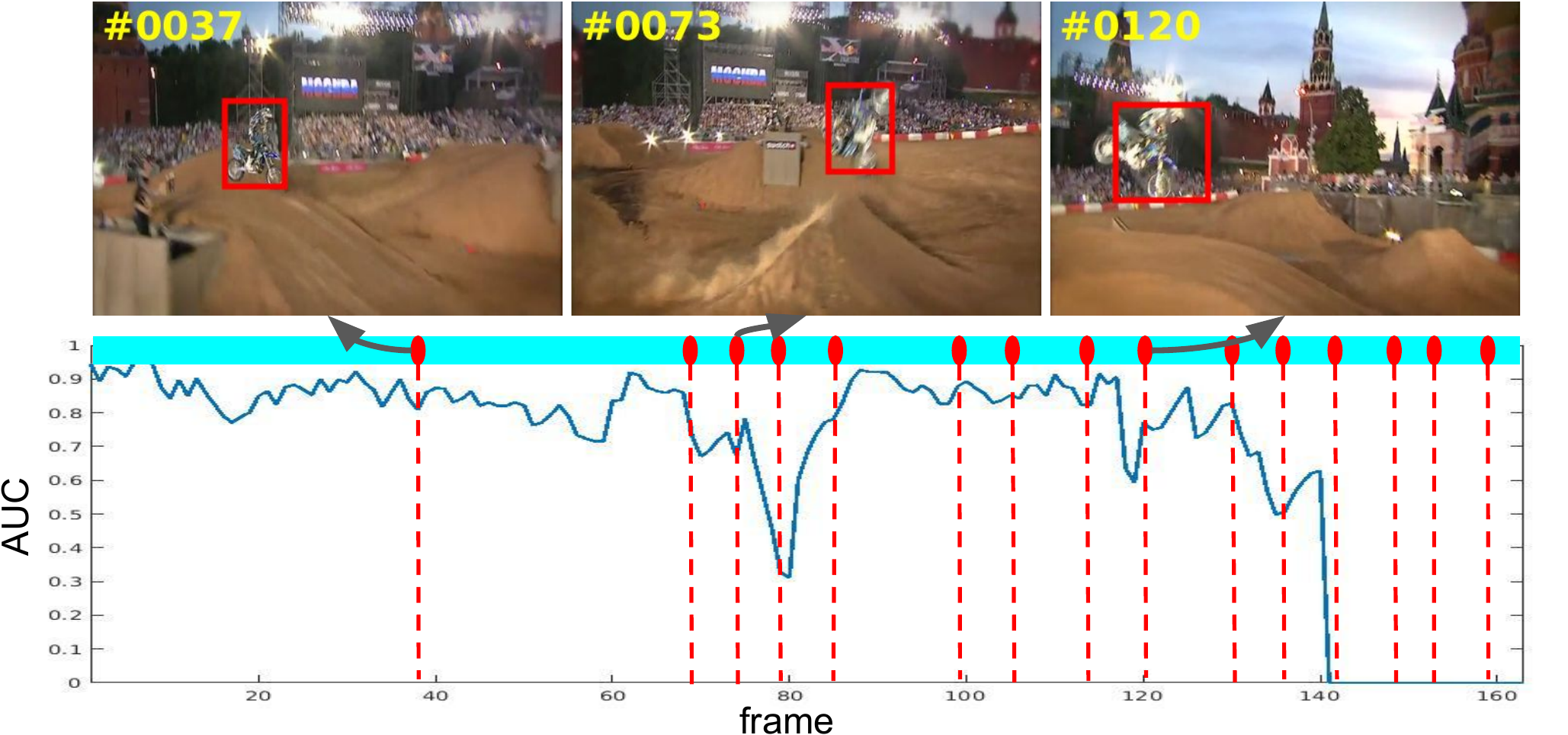}
\includegraphics[trim={0mm 45mm 45mm 0mm},clip,width=0.49\linewidth]{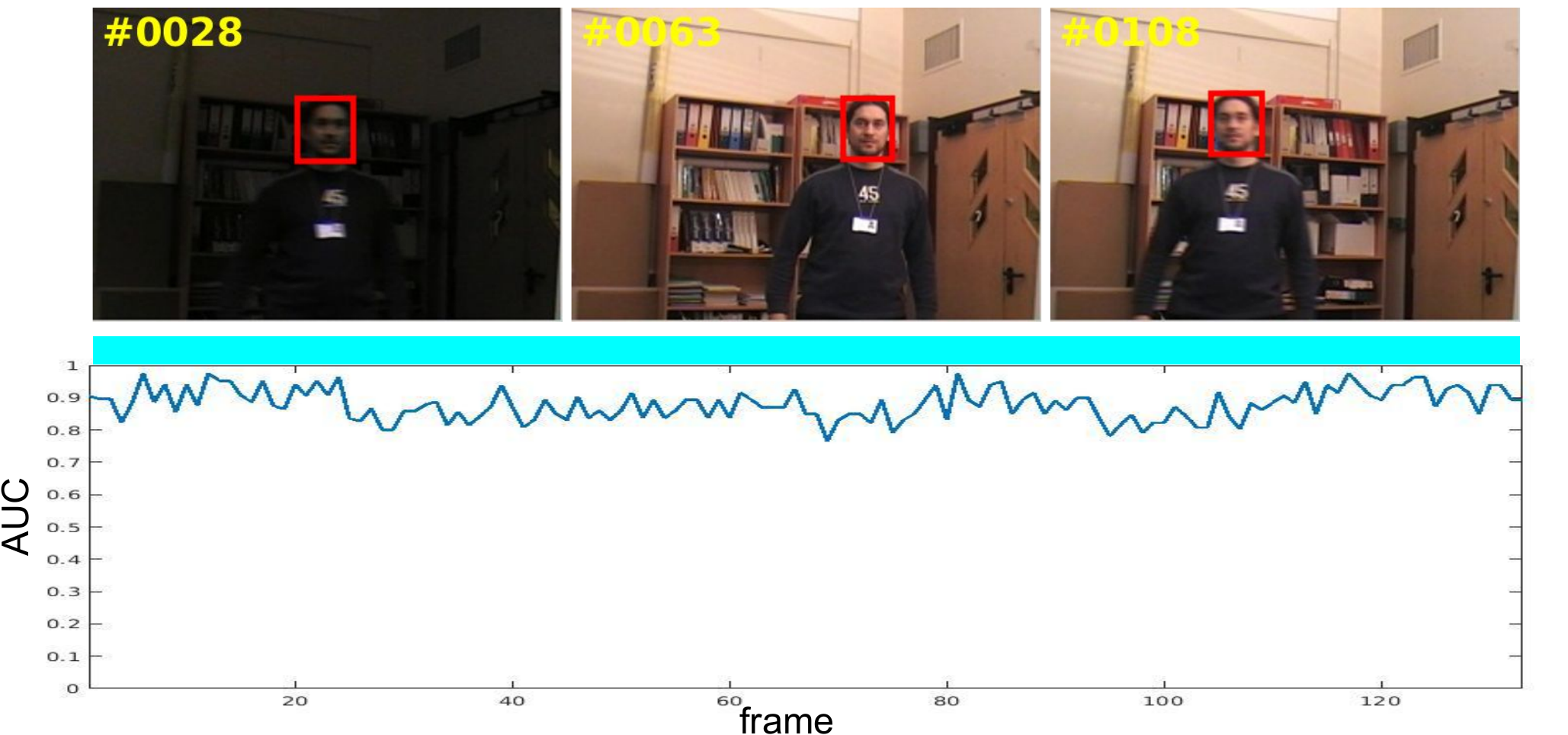}
\end{center}
\caption{Illustration of the detected failure alarms during online tracking by our AFAT in sequences \textit{DragonBoy}, \textit{Couple}, \textit{Bolt2}, \textit{Human9}, \textit{Gym}, \textit{Jogging2}, \textit{MotorRolling}, and \textit{Man}~\cite{wu2015object}. Red ovals denote the detected failure alarms by the proposed QPN.}
\label{fig8}
\end{figure}

\section{Evaluation}\label{experiment}
\subsection{Implementation details and experimental settings}
\label{sec_setting}
We implement our AFAT using PyTorch 1.0.1 on a platform with an Intel(R) Xeon(R) Gold 6134 CPU and three NVIDIA GeForce RTX 2080Ti GPU cards. For speed test, we only use a single GPU card. The code will be made publicly available.
We set $K=20$ for temporal sequential data collection.
We use the same parameters as SiameseRPN++~\cite{li2019siamrpn++} for the backbone networks, mobilev2 and resnet50, released by the original authors for all the experiments (no additional optimised parameters are used).
We compare the proposed AFAT on several benchmarks, including LaSOT~\cite{fan2019lasot}, UAV123~\cite{mueller2016benchmark}, OTB2015~\cite{wu2015object}, VOT2016~\cite{Kristan2016The}, VOT2018~\cite{kristan2018sixth} and VOT2019~\cite{kristan2019seventh}, with a number of state-of-the-art trackers, \textit{i.e.}, ATOM~\cite{danelljan2019atom}, Meta-Tracker~\cite{park2018meta}, SiamRPN++~\cite{li2019siamrpn++},  DaSiam~\cite{zhu2018distractor}, SiamRPN~\cite{li2018high}, LADCF~\cite{xu2019learning}, CREST~\cite{song2017crest},  BACF~\cite{kiani2017learning}, CFNet~\cite{valmadre2017end},  CACF~\cite{mueller2017context},  CSR-DCF~\cite{lukezic2017discriminative},  C-COT~\cite{danelljan2016beyond}, Staple~\cite{bertinetto2016staple}, SiameseFC~\cite{bertinetto2016fully},  SRDCF~\cite{danelljan2015learning} and other top-ranking trackers in VOT challenges.

To measure the tracking performance, we follow the corresponding protocols~\cite{wu2015object,kristan2015visual,kristan2016novel}.
We use precision plot and success plot~\cite{wu2013online} for  OTB2015, UAV123 and LaSOT.
Specifically, the precision plot measures the percentage of frames with the distance between the tracking results and ground truth less than a certain number of pixels.
The success plot measures the proportion of successful frames with the threshold ranging from 0 to 1 (a result is considered successful if the overlap of the predicted and ground-truth bounding boxes exceeds a pre-defined threshold).
Three numerical values, \textit{i.e.} distance precision (\textbf{DP}), normalised precision (\textbf{NP}) and area under curve (\textbf{AUC}), are further employed to measure the performance.
\textbf{DP} is the corresponding precision plot value (illustrated in the legend of precision plot) when the threshold set to 20 pixels.
\textbf{AUC} is the expected success rate (illustrated in the legend of success plot) in terms of overlap evaluation.
For all the VOT (2016, 2018 and 2019) datasets, we employ expected average overlap (\textbf{EAO}), accuracy (\textbf{A}) value and robustness (\textbf{R}) to evaluate the performance~\cite{kristan2015visual}.

\begin{table*}[t]
\footnotesize
\renewcommand{\arraystretch}{1.1}
\caption{Tracking results on VOT2018. (The best three results are highlighted by {\color{red}{red}}, {\color{blue}{blue}} and {\color{brown}{brown}}.)}
\label{vot2018}
\centering
\begin{tabular}{lccccccc}
\toprule[2pt]
\multirow{2}{*}{}& ECO& CFCF& CFWCR & LSART & UPDT & SiamRPN  & LADCF \\
& \cite{danelljan2017eco} &\cite{gundogdu2018good}&\cite{he2017correlation}&\cite{sun2018learning} & \cite{bhat2018unveiling} & \cite{li2018high}  & \cite{xu2019learning} \\
\hline
\hline
\textbf{EAO} & 0.280 & 0.286 & 0.303 & 0.323  & 0.378 & 0.383  & 0.389 \\
\textbf{A} & 0.483 & 0.509 & 0.484 & 0.493 & 0.536 & 0.586  & 0.503\\
\textbf{R} & 0.276 & 0.281 & 0.267 & 0.218 & {\color{brown}{\textbf{0.184}}} & 0.276 & {\color{blue}{\textbf{0.159}}} \\
\toprule[2pt]
\multirow{2}{*}{}& Gradnet & DaSiamRPN & GFS-DCF & ATOM & SiamRPN++&SiamRPN++ & \textbf{AFAT}\\
&\cite{li2019gradnet}&\cite{zhu2018distractor}&  \cite{xu2019joint} & \cite{danelljan2019atom} & mobilev2 &resnet50 &\\
\hline
\hline
\textbf{EAO} &0.247&0.326&0.397 & 0.401 &{\color{brown}{\textbf{0.410}}}& {\color{blue}{\textbf{0.414}}}& {\color{red}{\textbf{0.419}}}\\
\textbf{A} & 0.507&0.56&0.511 & {\color{brown}{\textbf{0.590}}} &0.586& {\color{blue}{\textbf{0.600}}} & {\color{red}{\textbf{0.605}}}\\
\textbf{R} & 0.375& 0.34&{\color{red}{\textbf{0.143}}} & 0.204 &0.229& 0.234 & 0.239\\
\toprule[2pt]
\end{tabular}
\end{table*}

\begin{figure}[!]
\begin{center}
\includegraphics[trim={30mm 70mm 30mm 70mm},clip,width=0.49\linewidth]{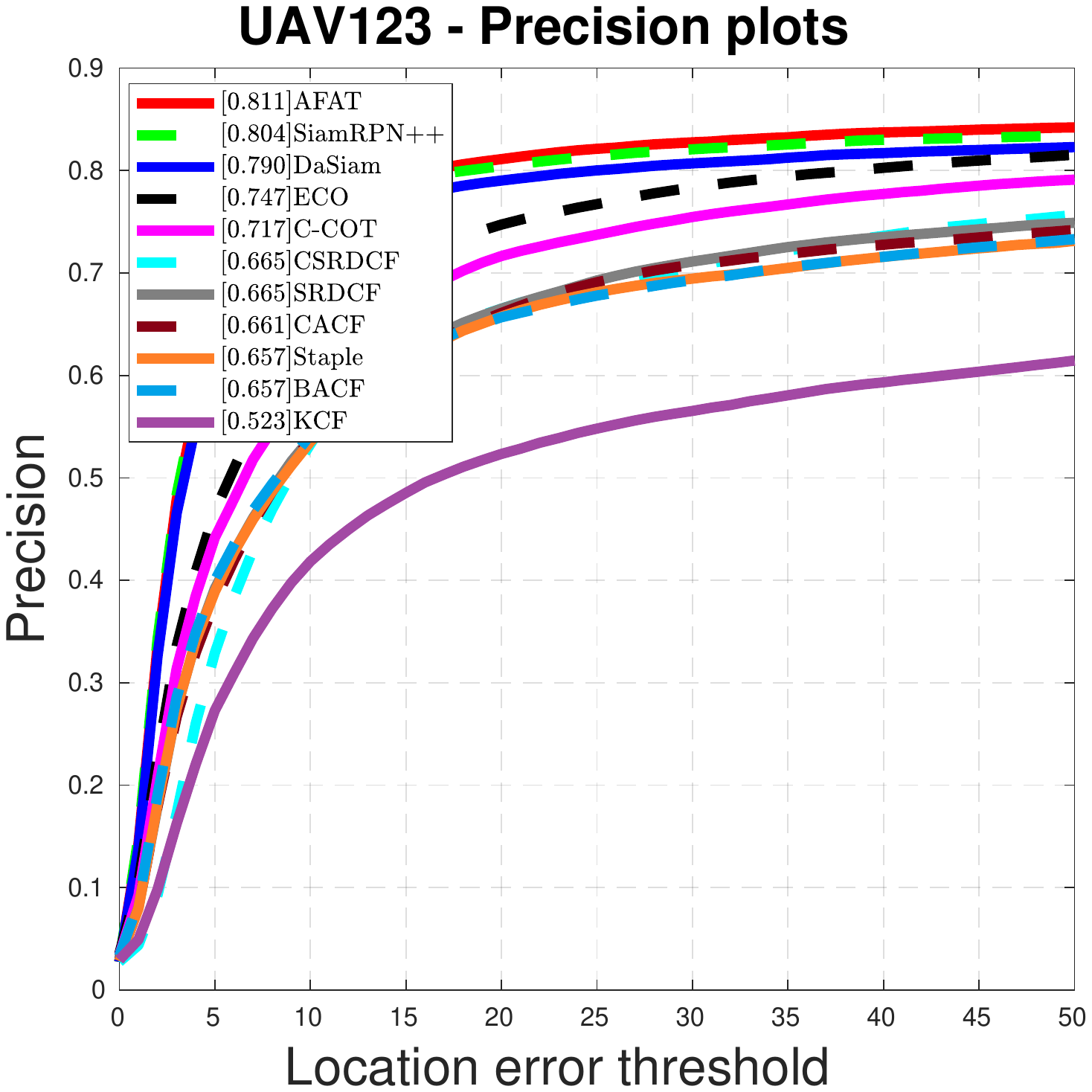}
\includegraphics[trim={30mm 70mm 30mm 70mm},clip,width=0.49\linewidth]{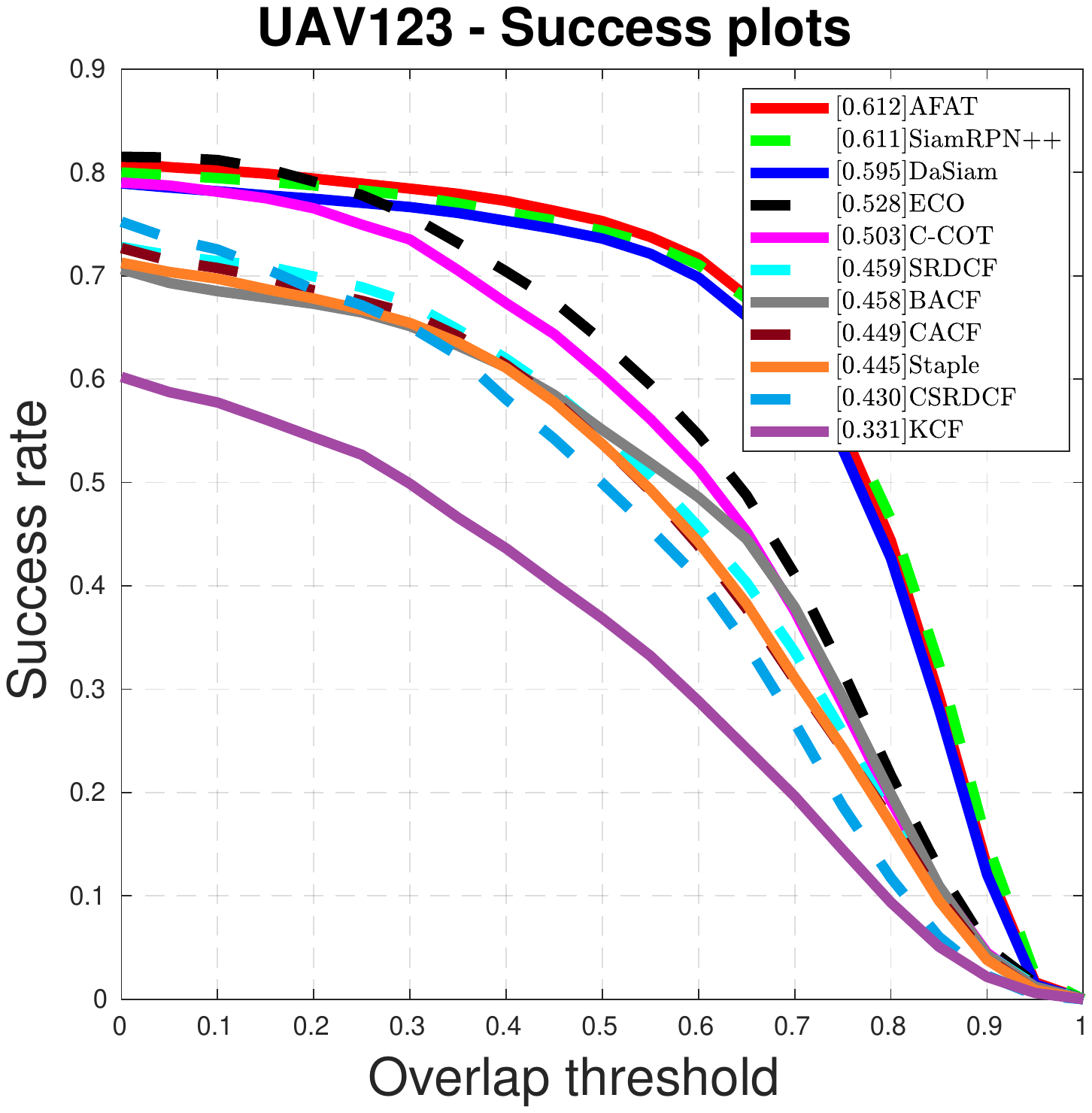}
\end{center}
\caption{The experimental results on UAV123. The precision plots with \textbf{DP} reported in the figure legend (\textit{left}) and the success plots with \textbf{AUC} reported in the figure legend (\textit{right}) are presented.}
\label{uav}
\end{figure}

\subsection{Ablation study}
To validate the effectiveness of the proposed AFAT, we perform ablation studies on recent public datasets including VOT2016, VOT2018, VOT2019, OTB2015, UAV123 and LaSOT.
As reported in Table~\ref{aba}, the proposed AFAT achieves improved performance compared to the base tracker, SiamRPN++$_\textrm{mobilev2}$, with light sacrifice on speed.
Specifically, though SiamRPN++$_\textrm{resnet50}$ is employed as a correction tracker in our design, AFAT outperforms SiamRPN++$_\textrm{resnet50}$ on \textbf{EAO} and \textbf{A} on all the three VOT datasets.
Due to the failure-aware system, some challenging frames are recognised by our QPN with follow-up corrections, alleviating the possibility of being judged as failures under VOT supervised evaluation.
The overall performance of SiamRPN++$_\textrm{resnet50}$ is ahead of SiamRPN++$_\textrm{mobilev2}$ on VOT2016 and VOT2018, while shows the opposite on VOT2019.
However, the detailed accuracy of these two trackers is diverse for each individual sequence, supporting the conclusion that the correction tracker is not necessarily stronger than the base tracker.
In addition, the \textbf{AUC} values on OTB2015 and UAV123 of AFAT are also better than SiamRPN++$_\textrm{resnet50}$, demonstrating the advantage of the proposed quality prediction network under unsupervised evaluation.
But AFAT cannot outperform SiamRPN++$_\textrm{resnet50}$ on LaSOT with \textbf{AUC} metric, as LaSOT dataset shares a longer average sequence length than others, with more demands on long-term robustness.

The failure detection numbers are also reported for each dataset in Table~\ref{aba}.
It is interesting that our AFAT gives more failure detection results on the VOT datasets, with the detection rate ranging from 30.9\% to 34.2\%.
While our QPN detects more successful results on OTB2015, UAV123 and LaSOT, which is consistent with the insight that VOT datasets generates more difficulties for shot-term visual tracking.
In addition, we illustrate the detected failure alarms during online tracking in Fig.~\ref{fig8}.
Though there are false alarms during the tracking procedure, many positive alarms are detected by our QPN with a conservative detection strategy in the training stage.
With further correction, a number of potential failures can be corrected to a certain degree.
The above results and analysis demonstrate the efficiency of AFAT for failure detection.

\begin{table}[tbp]
\footnotesize
\renewcommand{\arraystretch}{1.1}
\caption{Tracking results on VOT2019. (The best three results are highlighted by {\color{red}{red}}, {\color{blue}{blue}} and {\color{brown}{brown}}.)}
\label{vot2019}
\centering
\begin{tabular}{lccccccc}
\toprule[2pt]
\multirow{2}{*}{}& SiamRPN & SiamMask & GFS-DCF & ATOM & SiamRPN++&SiamRPN++ & \textbf{AFAT}\\
&\cite{li2018high}&\cite{wang2019fast}&  \cite{xu2019joint} & \cite{danelljan2019atom} & mobilev2 &resnet50 &\\
\hline
\hline
\textbf{EAO} &0.0.285&0.287&0.291 & {\color{blue}{\textbf{0.292}}} &{\color{blue}{\textbf{0.292}}}& 0.287& {\color{red}{\textbf{0.295}}}\\
\textbf{A} & {\color{blue}{\textbf{0.599}}}&0.594&0.513 & {\color{red}{\textbf{0.603}}} &0.580& 0.595 & {\color{blue}{\textbf{0.599}}}\\
\textbf{R} & 0.482& 0.461&0.453 & {\color{red}{\textbf{0.411}}} &{\color{blue}{\textbf{0.446}}}& 0.467 & {\color{brown}{\textbf{0.450}}}\\
\toprule[2pt]
\end{tabular}
\end{table}
\begin{figure}[t]
\begin{center}
\includegraphics[trim={0mm 48mm 0mm 35mm},clip,width=0.49\linewidth]{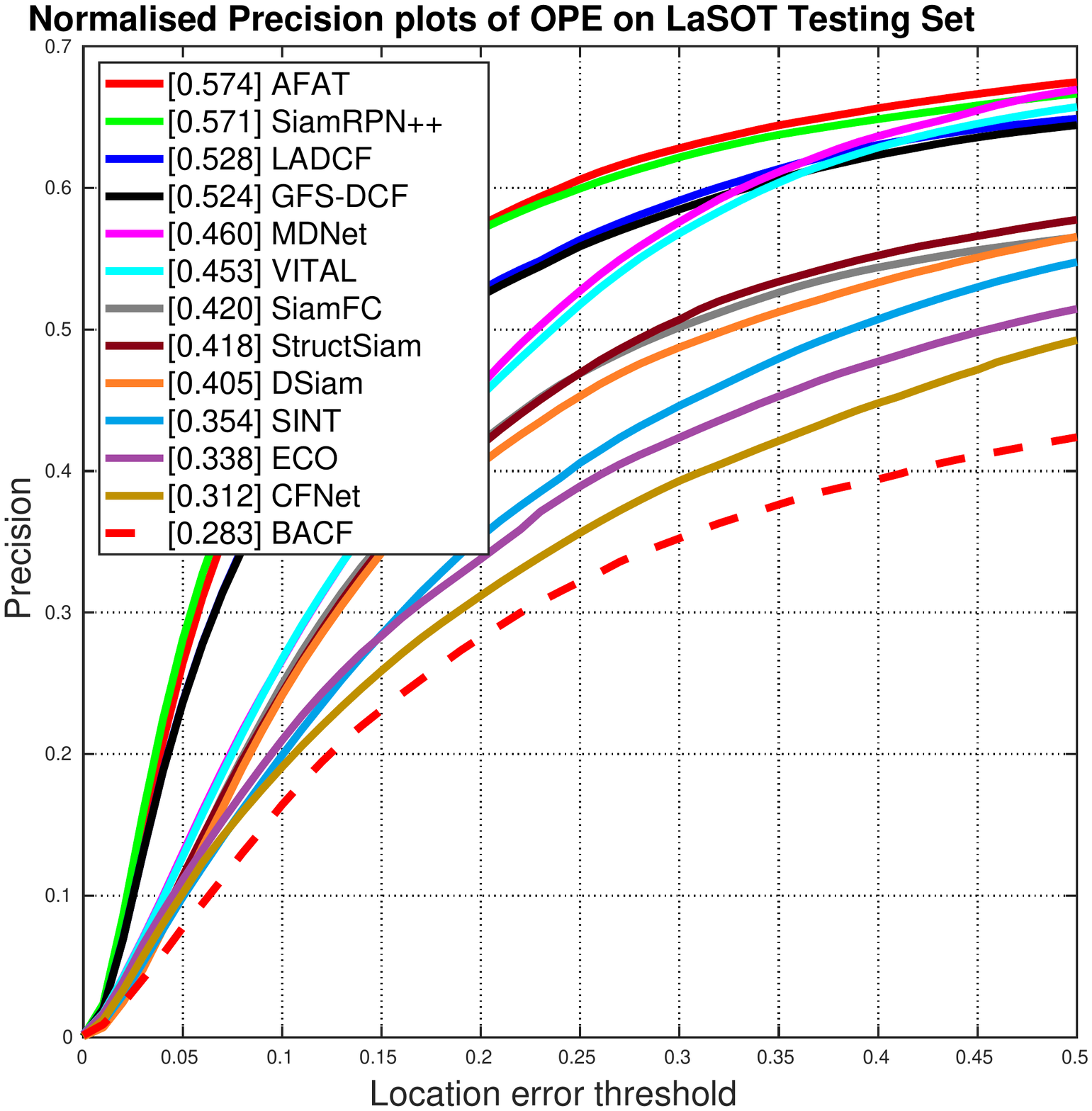}
\includegraphics[trim={0mm 48mm 0mm 35mm},clip,width=0.49\linewidth]{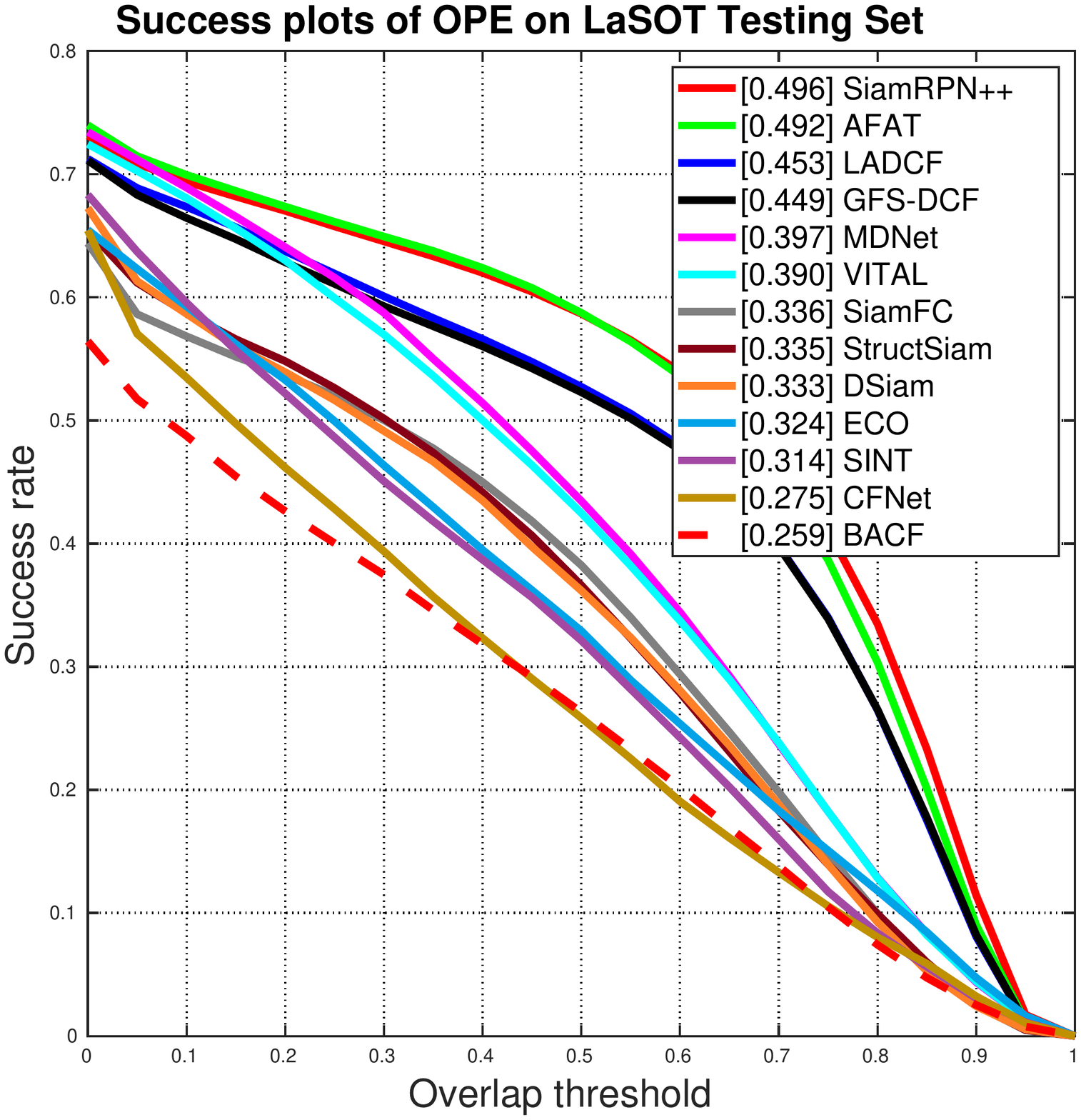}
\end{center}
\caption{The experimental results on LaSOT. The normalised precision plots with \textbf{NP} reported in the figure legend (\textit{left}) and the success plots with \textbf{AUC} reported in the figure legend (\textit{right}) are presented.}
\label{lasot}
\end{figure}

\subsection{Comparison with the state-of-the-art}
\noindent\textbf{The VOT2018 Dataset:} We first evaluate our AFAT on the recent VOT2018 dataset~\cite{kristan2018sixth} and compare the tracking performance with 13 state-of-the-art trackers.
The VOT2018 public dataset contains 60 challenging video sequences for single object tracking research.
Following the standard protocol~\cite{kristan2016novel}, we report the detailed results in Table~\ref{vot2018}.
AFAT achieves the best \textbf{EAO} score, 0.419, and \textbf{Accuracy}, 0.605, outperforming recent advanced trackers,~\textit{e.g.}, LADCF, ATOM and SiamRPN++.
In addition, the reported speed of AFAT on VOT2018 is 70.5 FPS (Table~\ref{vot2018}), demonstrating the effectiveness and efficiency of the proposed adaptive failure-aware system.

\noindent\textbf{The VOT2019 Dataset:} We also evaluate our AFAT on the VOT2019 dataset~\cite{kristan2019seventh} and compare the tracking performance with 6 state-of-the-art trackers.
The VOT2019 public dataset contains 60 challenging video sequences for single object tracking research, introducing more challenging factors than VOT2018.
Following the standard protocol~\cite{kristan2016novel}, we report the detailed results in Table~\ref{vot2019}.
According to the table, the proposed AFAT method achieves the best \textbf{EAO} score, 0.295, outperforming recent advanced trackers,~\textit{e.g.}, ATOM (0.292) and SiamRPN++ (0.292).
In addition, the reported \textbf{Accuracy} (0.599) and \textbf{Robustness} (0.450) scores of AFAT are also within the top three, demonstrating the effectiveness of the proposed adaptive failure detection and correction framework.

\noindent\textbf{The UAV123 Dataset:} UAV123 dataset contains 123 video sequences, recorded by a camera embedded in a UAV system~\cite{mueller2016benchmark}.
The average sequence length of UAV123 is 915 frames.
We report the experimental results in Fig.~\ref{uav} with precision and success plots.
Our AFAT outperforms recent DCF and Siamese trackers: ECO, CSRDCF, DaSiam and SiamRPN++.
Specifically, the proposed AFAT method achieves improved performance in terms of \textbf{AUC} (0.612), \textbf{DP} (0.811) and speed (80.4 FPS), against SiamRPN++ with \textbf{AUC} (0.611), \textbf{DP} (0.804) and speed (37.9 FPS).

\noindent \textbf{The LaSOT Dataset:} LaSOT is a recent dataset including 1400 sequences in 70 object categories, with an average sequence length of more than 2500 frames.
We evaluate the tracking performance on the designed LaSOT test set that contains 280 video sequences, with 4 videos of each category~\cite{fan2019lasot}.
Fig.~\ref{lasot} reports the normalised precision plots and success plots of the proposed AFAT and recent state-of-the-art trackers, \textit{e.g.}, LADCF, MDNet, VITAL, and SiamRPN++.
AFAT achieves 0.574 in the \textbf{NP} metric, outperforming SiamRPN++ (0.571).
But SiamRPN++ achieves better performance in terms of \textbf{AUC} than AFAT.

To summarise, our AFAT method achieves advanced performance, as compared with the state-of-the-art trackers.
In addition, the speed of our AFAT is ranging from 70 FPS to 87 FPS, depending on the number of detected failures.
It should be highlighted that a forward pass of our QPN takes only 1.2$\sim$1.5 ms, demonstrating its feasibility in real-time visual object tracking.

\section{Conclusion}\label{conclusion}
We proposed an effective online Quality Prediction Network (QPN) by learning sequential response maps in the decision level, enabling adaptive quality perception of an online visual object tracker.
With spatial feature extraction and temporal feature fusion, the quality of the current tracking result can be trained from sequential response maps in a supervised manner, achieving online failure detection.
Furthermore, we combined our QPN with the recent SiamRPN++ method to construct our AFAT tracker.
The significance of realising online failure detection and correction was examined in datasets with diverse challenges.
The extensive experimental results obtained on a number of benchmarking datasets demonstrate the effectiveness and robustness of the proposed method, as compared with the state-of-the-art trackers.

%
%
\bibliographystyle{splncs04}
\bibliography{main}
\end{document}